\documentclass{article}
\usepackage{ijcai20}

\usepackage{times}

\usepackage{soul}
\usepackage{url}
\usepackage[hidelinks]{hyperref}
\usepackage[utf8]{inputenc}
\usepackage[small]{caption}
\usepackage{graphicx}
\usepackage{amsmath}
\usepackage{booktabs}
\usepackage{amsmath}
\usepackage{amssymb}
\usepackage{mathrsfs}
\usepackage{txfonts}
\usepackage{bm}
\usepackage{multirow}
\usepackage{tabularx}
\usepackage{caption,subcaption}
\usepackage{booktabs}
\usepackage{algorithm}
\usepackage{algorithmic}
\usepackage{float}
\usepackage{array}
\usepackage{makecell}

\title{Self-adaptive Re-weighted Adversarial Domain Adaptation}

\author{
Shanshan Wang\and
Lei Zhang\footnote{Contact Author}
\\
\affiliations
Learning Intelligence $\&$ Vision Essential (LiVE) Group\\
School of Microelectronics and Communication Engineering, Chongqing University, Chongqing, China\\
\emails
\{wangshanshan, leizhang\}@cqu.edu.cn
}

\begin{document}

\maketitle

\begin{abstract}
Existing adversarial domain adaptation methods mainly consider the marginal distribution and these methods may lead to either under transfer or negative transfer. To address this problem, we present a self-adaptive re-weighted adversarial domain adaptation approach, which tries to enhance domain alignment from the perspective of conditional distribution. In order to promote positive transfer and combat negative transfer, we reduce the weight of the adversarial loss for aligned features while increasing the adversarial force for those poorly aligned measured by the conditional entropy.
Additionally, triplet loss leveraging source samples and pseudo-labeled target samples is employed on the confusing domain. Such metric loss ensures the distance of the intra-class sample pairs closer than the inter-class pairs to achieve the class-level alignment.
In this way, the high accurate pseudo-labeled target samples and semantic alignment can be captured simultaneously in the co-training process.  Our method achieved low joint error of the ideal source and target hypothesis. The expected target error can then be upper bounded following Ben-David's theorem.
Empirical evidence demonstrates that the proposed model outperforms state of the arts on standard domain adaptation datasets.
\end{abstract}

\section{Introduction}  


 \begin{figure}[h]
\begin{center}
 \includegraphics[width=1.0\linewidth]{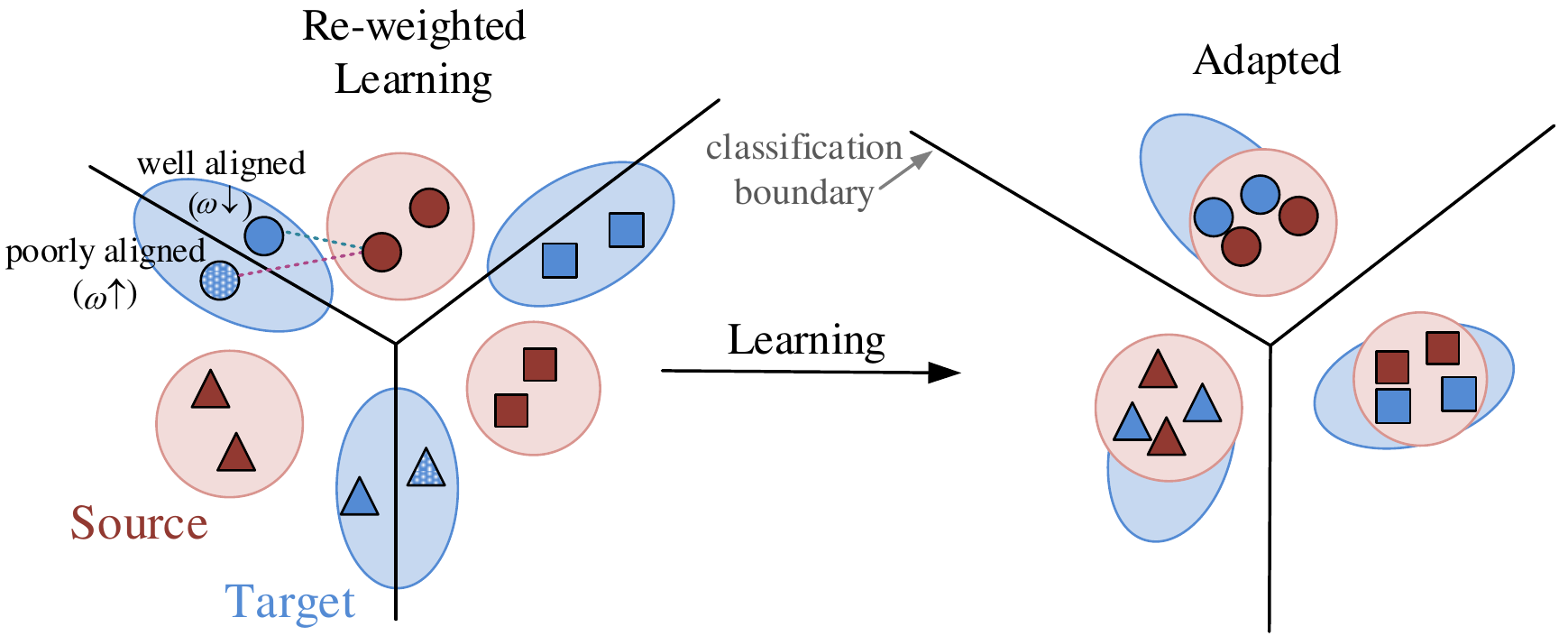}
\end{center}
  \setlength{\abovecaptionskip}{0pt}
   \caption{Motivation of our method. Let $\omega$ express weight, then down-weight~($\omega\downarrow$) well aligned samples and up-weight~($\omega\uparrow$) poorly aligned samples according to uncertainty. Different shapes represent different classes.  Different colors mean different domains. Shadow denotes the misclassified samples.}
\label{fig0}
\end{figure}

Unsupervised Domain Adaptation (UDA) ~\cite{pan2010survey} task aims to recognize the unlabeled target domain data, leveraging a sufficiently labeled, related but different source domain. The key issue of UDA is to reduce distribution difference between the two domains, such that the learned classifier from source domain can well classify target domain samples.
Generally, maximum mean discrepancy (MMD)~\cite{long2015learning}, as a non-parametric metric, is commonly used to measure the dissimilarity of distributions.
Recently, adversarial learning~\cite{bousmalis2016domain} has been successfully brought into UDA to reduce distribution discrepancy, in which domain-invariant or domain-confused feature representation is usually learned.
Unlike many previous MMD-based methods, domain-adversarial neural networks focus on combining UDA and deep feature learning within a unified training paradigm.
The goal of adversarial domain adaptation is to confuse the features between domains, so that domain-invariant representations are ultimately obtained.


However, as discussed in MCD~\cite{saito2017maximum}, it does not really guarantee safe domain alignment. \textit{i.e.}, the alignment of category space between domains is ignored in alleviating domain shift. Target samples that are close to the decision boundary or far from their class centers could be misclassified by the classifier trained in source domain~\cite{wen2016discriminative}. Thus, previous domain adversarial adaptation methods that only match the domain distributions without exploiting the inner structures may be prone to under transfer~(underfitting) or negative transfer~(overfitting).

To address this problem, ~\cite{SaitoATDA2017} attempt to include the target samples into the learning of their models. Specifically, some method~\cite{PFAN2019} proposed to leverage pseudo-labels which is progressively guaranteed by an Easy-to-Hard strategy to learn target discriminative representations. These methods encourage a low-density separation between classes in the target domain.

However, as the domain bias exists,  the pseudo labeled samples are not always correct. In order to obtain the high accurate labels of target samples, a much closer domain distribution is expected as the source classifier can generalize well on such domain-invariant target representations.
In this paper, to tackle the aforementioned challenge, we take a two-step approach to learn domain invariant representations.

Firstly, the structure of adversarial network is  adopted in our method. Although existing adversarial learning methods aim to reduce domain distribution discrepancy, they still suffer from a major limitation: these approaches mainly consider the marginal distribution while ignoring the conditional distribution, the classifier learned from source domain might be incapable of confidently distinguishing target samples. \textit{i.e.}, the joint distributions of feature and category are not well aligned across domains.
Thus, these methods  may lead to either under transfer or negative transfer.

To promote positive transfer and combat negative transfer, we propose to recognize the transferability of each sample and re-weight these samples to force the underlying domain distributions closer. To push further along this line, we propose a self-adapted re-weighted adversarial DA approach shown in Fig.~\ref{fig0}. Our method considers the transferable degree from the perspective of conditional distribution, therefore it can adapt better on target domain than previous approaches which only consider marginal distribution.

In information theory, the entropy is an uncertainty measure which can also be borrowed to quantify the adaptation.
Different from previous methods which employ the conditional entropy directly, we utilize the entropy criterion to generate the weight and measure the degree of domain adaptation. Noteworthily, the conditional entropy is constructed
by the conditional distribution. \textit{i.e.}, our model does not just reduce conditional distributions between two domains directly, but dynamically leverages the conditional distribution to re-weight the samples.
The inner reason lies that if the sample can get a high prediction by the conditional entropy, it can be regarded as a poorly-aligned sample, otherwise a well-aligned sample.
The weights for those well aligned features are decreasing while for those poorly aligned features increasing in adversarial loss self-adaptively, then a better domain-level alignment can be achieved. If the distribution bias is reduced, the precisely pseudo-labeled samples in target domain can be chosen.

Secondly, as the pseudo labels are not always correct, they are not directly leveraged to train the classifier. Instead, they are  employed to train the generalized feature representations.
In our method, not only global domain-level aligning strategy, but also the metric loss is employed to learn the discriminative distance in the confusing domain. In our mechanism, triplet loss utilizes source samples and pseudo-labeled target samples to keep the samples align well in class-level.
As a result, our model can learn better domain alignment features in the collaborative alignment adversarial training process and these features are not only domain invariant but also class discriminative for semantic alignment.
This will have a more confident guarantee that the joint error of the ideal source and target hypothesis is low. The DA becomes possible as presented in Ben-David's theorem~\cite{Ben2010A}.

The main contributions and novelties of this paper are summarized as follows.
\begin{itemize}
\item Our model attempts to learn the target generalized model to promote positive transfer and combat
  negative transfer. The network leveraging the joint distributions is much more proper than only the marginal distribution.
\item In order to achieve the better domain confusion, we present a self-adaptive re-weighted adversarial domain adaptation approach through entropy from the perspective of conditional distribution.In our method, the adversarial network forces to reduce domain discrepancy by re-weighting the samples. The weights of the adversarial loss for well aligned features are decreased while increasing for those poorly aligned, such that a better domain-level alignment can be achieved.
\item Besides the domain-level alignment, triplet loss is employed to enforce the features have better inter-class separation and intra-class compactness utilizing source samples and pseudo-labeled target samples. As a result, the feature representations which are not only domain invariant for domain alignment but also class discriminative for semantic alignment can be learned.
\end{itemize}

\section{Related Work} 

Training CNN for UDA can be conducted through various strategies. Matching distributions of the middle features~\cite{long2015learning,long2017deep,zellinger2017central} in CNN is considered to be effective for an accurate adaptation. These works pay attention to first-order or high-order statistics alignment.

Recent research on deep domain adaptation further embeds domain-adaptation modules in deep networks to boost transfer performance. In~\cite{ganin2017domain}, DANN is proposed for domain adversarial learning, in which a gradient reversal layer is designed for confusing features from two domains. This method can be regarded as the baseline of adversarial learning methods.
Tzeng \textit{etal.} proposed an ADDA method~\cite{Tzeng2017Adversarial} which combines discriminative modeling, untied weight sharing and a GAN loss.
Long \textit{etal.}~\cite{Long2017Conditional} also present a conditional adversarial domain adaptation (CDAN) that conditions the discriminative information conveyed into the predictions of classifier.
Wang et al.~\cite{TADA2019} propose a TADA focus the adaptation model on transferable regions or images both by local attention and  global attention.

However, these methods are based on the theory that the predicted error is bounded by the distribution divergence. They do not consider the relationship between target samples and decision boundaries. To tackle these problems, multiple classifiers instead of the discriminator are considered and these methods become another branch.
Saito et al. propose a ATDA method~\cite{SaitoATDA2017} by tri-training three classifiers equally to give pseudo-labels to unlabeled samples.
Later, he also proposes a new approach called MCD~\cite{saito2017maximum} that uses two different classifiers to align those easily misclassified target samples through adversarial learning in CNN. Recently, Zhang et al. proposed SymNet ~\cite{Zhang2019Domain} based on a symmetric design of source and target task classifiers, meanwhile, an additional classifier that shares with layer neurons was constructed.

In our method, we propose a different strategy to address these problems. Not only the source data but also the target samples are leveraged to align domain features and class relations.

 \begin{figure}
\begin{center}
 \includegraphics[width=0.9\linewidth]{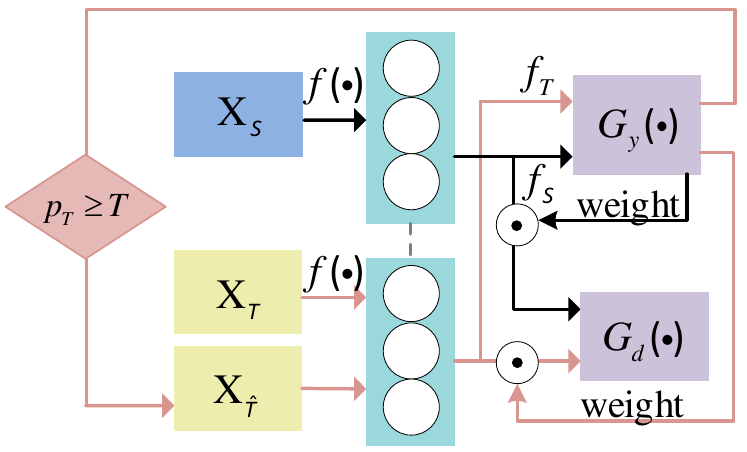}
\end{center}
  \setlength{\abovecaptionskip}{0pt}
   \caption{The framework of our method. To promote positive transfer and combat negative transfer, we utilize the entropy criterion to reveal the transferable degree of samples, then re-weight them and feed them into the discriminative network to force the underlying distributions closer.
   }
\label{fig1}
\end{figure}
\section{Self-adaptive Re-weighted Adversarial DA}
An overview of our method is depicted in Fig.~\ref{fig1}. In UDA, we suppose $\mathcal{D}_s=\{(x_i^s,y_i^s)\}_{i=1}^{n_{s}}$ and $\mathcal{D}_t=\{x_j^t\}_{j=1}^{n_{t}}$ to be the labeled source data and unlabeled target data, drawn from different distributions respectively. Our goal is to predict the target label $\hat{y}^t = \arg \max {G_y}(f(x^t))$ and minimize the target risk $\epsilon_t(G_y)=\mathbb{E}_{(x^t,y^t)\sim {\mathcal{D}_t}}[G_y (f(x^t))\neq y^t]$, where $G_y(\cdot)$ represents the softmax output and $ f(\cdot) $ refers to the feature representation.

Our method aims to construct a target generalized network. The re-weighted adversarial domain adaptation forces a close global domain-level alignment. Simultaneously, triplet loss is leveraged to train the class-discriminative representations utilizing source samples and pseudo-labeled samples. Then the classifier gradually increase accuracies on the target domain.

\textbf{Preliminaries: Domain Adversarial Network.}
In DA setting, domain adversarial networks have been successfully explored to minimize the cross-domain discrepancy by extracting transferable features. The procedure is a two-player game: the first player is the domain discriminator $G_d$ trained to distinguish the source domain from the target domain, and the second player is the feature extractor $f(\cdot)$ trained to confuse the domain discriminator. The objective function of domain adversarial network is as follows:
\setlength\abovedisplayskip{2pt}
\setlength\belowdisplayskip{2pt}
\begin{equation}\small
\begin{split}
 {\mathcal{L}}^{s}_{task} (\theta _f,\theta _y )& = \frac{1}{{n_s }}\sum\limits_{{\bf{x}}_i \in \mathcal{D}_s } {{\mathcal{L}}_y } (G_y (f ({\bf{x}}_i )),{\bf{y}}_i^s ), \\
 {\mathcal{L}}_D (\theta _f ,\theta _y ,\theta _d )& = -\frac{1 }{n_s+n_t}\sum\limits_{{\bf{x}}_i  \in (\mathcal{D}_s  \cup \mathcal{D}_t )} {{\mathcal{L}}_d } (G_d (f ({\bf{x}}_i )),d_i ),\\
\end{split}
\label{eequa1}
\end{equation}
where $\theta_f, \theta_d$ represent the parameters of feature network and domain discriminator, respectively. $\theta_y$ is the parameter of source classifier. $d_i$ is the domain label of sample $x_i$.

\textbf{Self-adaptive Re-weighted Adversarial DA.}
In practical domain adaptation problems, however, the data distributions of the source domain and target domain usually embody complex multimode structures. 
Thus, previous domain adversarial adaptation methods that only match the marginal distributions without exploiting the multimode structures maybe prone to either under transfer or negative transfer. To promote positive transfer and combat negative transfer, we should find a technology to reveal the transferable degree of samples and then re-weight them to force the underlying distribution closer.

\begin{figure}
\begin{center}
 \includegraphics[width=0.7\linewidth]{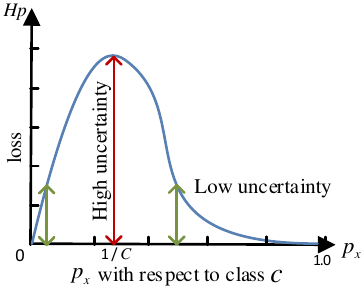}
\end{center}
  \setlength{\abovecaptionskip}{0pt}
   \caption{The conditional entropy measures the uncertainty. If the sample can get a low entropy, it can be regarded as a well-aligned transferable sample, else it is a poorly-aligned sample.}
\label{fig2}
\end{figure}

As mentioned earlier, not all images are equally transferable in domain adaptation network and some images are more transferable than others. Therefore, we propose a method to measure adaptable degree using the certainty estimate.  In information theory, the entropy functional is an uncertainty measure which nicely meets our need to quantify the adaptation and  is depicted in Fig.~\ref{fig2}.
Therefore, we utilize the entropy criterion to estimate weights and improve the domain confusion by relaxing the alignment on these well aligned samples and focusing the alignment on these poorly aligned.
If the sample can get a low entropy, it can be regarded as a well-aligned transferable sample, else it is a poorly aligned sample.
The conditional distribution leveraged in the entropy is not considered in the standard adversarial DA methods.
We adopt the conditional entropy as the indicator to weight the adversarial loss and the adversarial loss is extended as
\setlength\abovedisplayskip{2pt}
\setlength\belowdisplayskip{2pt}
\begin{equation}\small
\begin{split}
&{\mathcal{L}}_{adv} (\theta _f , \theta _d ) = -\frac{1}{n_s+n_t}\sum\limits_{{\bf{x}}_i  \in (\mathcal{D}_s  \cup \mathcal{D}_t )} ( 1+{{\mathcal{H}}_p}){{\mathcal{L}}_d } (G_d (f ({\bf{x}}_i )),d_i ),\\
& where ~~~~~{{\mathcal{H}}_p } =-\frac{1}{C}\sum\limits_{c = 1}^{C } p_c \log (p_c )\\
\end{split}
\label{eequa3}
\end{equation}
where $C$ is the number of classes and $p_c$ is the probability of predicting an sample to class $c$.

To make the best of conditional distribution, entropy minimization principle is adopted to enhance discrimination of learned models for target data by following ~\cite{long2016unsupervised}. In order to reduce wrong classified samples due to domain shift, the entropy minimization loss is used to update both the feature network and classifier.
\setlength\abovedisplayskip{2pt}
\setlength\belowdisplayskip{2pt}
\begin{equation}\small
\begin{split}
 {\mathcal{L}}_{h} (\theta _f,\theta _y ) = -\frac{1}{{n_t }}\sum\limits_{{\bf{x}}_i \in \mathcal{D}_t} p_t \log (p_t ).
\end{split}
\label{eequa5}
\end{equation}

\textbf{Class-level Alignment.}
So far, we only consider the global domain-level confusion, the discriminative power between classes is not involved~\cite{hong2015learning}. For one thing, samples with same labels should be pulled together in the embedding space. For another, samples with different labels should be separated apart. Naturally, metric learning~\cite{yang2017person} as an effective method can be implemented to achieve our goal. Triplet loss~\cite{Schroff2015FaceNet} tries to enforce a margin $m$ between each pair of samples from one class to all other classes. It allows samples to enforce the distance and thus discriminate to other classes. In our paper, we select triplet loss to train the class-level aligned features.

Intuitively, a target generalized classifier is much more proper than the source classifier for the target domain.
We propose to assign pseudo-labels to target samples and train the network as if they were true labels. Noteworthily, as the pseudo labels are not always correct, they are not directly leveraged to train the classifier.
In order to make the best of  pseudo-labeled samples, we leverage the source samples and pseudo-labeled target samples to construct the sample-pairs in metric learning. We follow the sampling strategy in~\cite{Deng2018Domain} to randomly select samples.

The pseudo label $\hat{y}^t_i$ of $x^t_i$ is predicted based on maximum posterior probability using the source cross-entropy loss. It is progressively updated during optimization. Additionally, we only select target images with predicted scores above a high threshold $T$ for building the semantic relations based on the intuitive consideration that the image with the high predicted score is more likely to be classified correctly. We empirically set the threshold $T$ as a constant.

In the confusing domain, given an anchor image ${\bf{x}}_a$, a positive image ${\bf{x}}_p$, and a negative image ${\bf{x}}_n$, the minimized loss is as:
\begin{equation}\small
\begin{split}
{\mathcal{L}}_{{\rm{tri}}} (\theta _f ) = \sum\limits_{\scriptstyle {\bf{x}}_i  \in (\mathcal{D}_s  \cup \mathcal{D}_t ) \hfill \atop
  \scriptstyle y_a  = y_p  \ne y_n  \hfill} {[m + d_{a,p}  - d_{a,n} ]_+}.
  \end{split}
\label{eequa4}
\end{equation}

\textbf{Overall Training Loss.}
With Eq.~(\ref{eequa1}), Eq.~(\ref{eequa3}), Eq.~(\ref{eequa5}) and Eq.~(\ref{eequa4}), the overall training loss of our model is given by,
\begin{equation}\small
\begin{split}
{\mathcal{L}}=\mathcal{L}^s_{task} +{\mathcal{L}}_{adv} +{\mathcal{L}}_{h} +{{\mathcal{L}}_{\rm{tri}}}.
\end{split}
\label{eequa6}
\end{equation}

The optimization problem is to find the parameters $\mathop {\theta _f }\limits^ \wedge,\mathop {\theta _y }\limits^ \wedge$ and $\mathop {\theta _d }\limits^ \wedge $ that jointly satisfy
\begin{equation}\small
\begin{array}{l}
 \mathop {(\theta _f }\limits^ \wedge  ,\mathop {\theta _y }\limits^ \wedge ) = \arg \mathop {\min }\limits_{\theta _f ,\theta _y } {\mathcal{L}} (\theta _f ,\theta _y ,\theta _d ) \\
 \mathop {(\theta _d }\limits^ \wedge) = \arg \mathop {\max }\limits_{\theta _d } {\mathcal{L}} (\theta _f ,\theta _y ,\theta _d). \\
 \end{array}
\label{eequa7}
\end{equation}


\section{Experiment}
In this section, several benchmark datasets, not only the toy datasets as USPS+MNIST datasets, but also Office-31 dataset~\cite{saenko2010adapting}, ImageCLEF-DA~\cite{long2017deep} dataset and Office-Home~\cite{venkateswara2017deep} dataset, are adopted for evaluation.

\textbf{Handwritten Digits Datasets}.
 USPS~(\textbf{U}) and MNIST~(\textbf{M}) datasets are toy datasets for domain adaptation. They are standard digit recognition datasets containing handwritten digits from $0-9$. USPS consists of 7,291 training images and 2,007 test images of size $16 \times 16$. MNIST consists of 60,000 training images
and 10,000 test images of size $28 \times 28$. We construct two tasks: $U \to M$ and $M \to U$ and we follow the experimental settings of~\cite{Hoffman2017CyCADA}.

\textbf{Office-31 Dataset}.
This dataset is a most popular benchmark dataset for cross-domain object recognition. The dataset consists of daily objects in an office environment and includes three domains such as Amazon (\textbf{A}), Webcam (\textbf{W}) and Dslr (\textbf{D}). There are 2,817 images in domain \textbf{A}, 795 images in \textbf{W} and 498 images in domain \textbf{D} making total 4,110 images.  With each domain worked as source and target alternatively, 6 cross-domain tasks are formed, \textit{e.g.}, \textbf{A} $\to$ \textbf{D} \textit{etc}.

\textbf{ImageCLEF-DA Dataset}.
The ImageCLEF-DA is a benchmark for ImageCLEF 2014 domain adaptation challenge. It contains 12 common categories shared by three public datasets: Caltech-256 (\textbf{C}), ImageNet ILSVRC 2012 (\textbf{I}) and Pascal VOC 2012 (\textbf{P}). In each domain, there are 50 images per class and totally 600 images are constructed. Images in ImageCLEF-DA are of equal size, making it a good alternative dataset. We evaluate all methods across three transfer domains and build 6 cross-domain tasks: \textit{e.g.}, \textbf{I} $\to$ \textbf{P} \textit{etc}.

\textbf{Office-Home Dataset}.
This is a new and challenging dataset for domain adaptation, which consists of 15,500 images from 65 categories coming from four significantly different domains: Artistic images (\textbf{Ar}), Clip Art (\textbf{Cl}), Product images (\textbf{Pr}) and Real-World images (\textbf{Rw}). With each domain worked as source and target alternatively, there are 12 DA tasks on this dataset.
The images of these domains have substantially different appearance and backgrounds, and the number of categories is much larger than that of Office-31 and ImageCLEF-DA, making it more difficult to transfer across domains.

\textbf{Results}.
In our experiment, the target labels are unseen by following the standard evaluation protocol of UDA~\cite{long2017deep}. Our implementation is based on the PyTorch framework. For the toy datasets of handwritten digits, we utilize the LeNet. For other datasets, we use the pre-trained ResNet-50 as backbone network.
We adopt the progressive training strategies as in CDAN~\cite{Long2017Conditional}. In the process of selecting pseudo-labeled samples, the threshold $T$ is empirically set as the constant 0.9. The margin $m$ and $N_0$ in triplet loss are set as 0.3 and 3 following the setting as usual, respectively.

We evaluate the rank-1 classification accuracy for comparison. For handwritten digits, as there are plenty of different configutations in these datsets, in order to give the fair comparison, we only show some the recent results with the same backbone and training/test split. we compared with  ADDA~\cite{Tzeng2017Adversarial}, CoGAN~\cite{liu2016coupled}, UNIT~\cite{Liu2017UNIT},  CYCADA~\cite{Hoffman2017CyCADA} and CDAN~\cite{Long2017Conditional}. For other datasets, our compared baseline methods include DAN~\cite{long2015learning}, DANN~\cite{ganin2017domain}, JAN~\cite{long2017deep}, CDAN~\cite{Long2017Conditional}and SAFN~\cite{xu2019larger}. Besides, on Office-31 dataset, we compare with TCA~\cite{pan2011domain}, GFK~\cite{Gong2012Geodesic}, DDC~\cite{tzeng2014deep}, RTN~\cite{long2016unsupervised}, ADDA~\cite{Tzeng2017Adversarial}, MADA~\cite{Cao2018Partial}, GTA~\cite{Sankaranarayanan2017Generate}, MCD~\cite{saito2017maximum}, iCAN~\cite{zhang2018collaborative}, TADA~\cite{TADA2019} and SymNet~\cite{Zhang2019Domain}. On ImageCLEF-DA dataset, RTN~\cite{long2016unsupervised}, MADA~\cite{Cao2018Partial} and iCAN~\cite{zhang2018collaborative} are compared.
On Office-Home dataset, TADA~\cite{TADA2019} and SymNet~\cite{Zhang2019Domain} are compared.

\begin{table}[t]\small
\begin{center}
\setlength{\tabcolsep}{1.5mm}{
\begin{tabular}{  l | c c   |c}
\toprule
Handwritten&M $\to$ U&U $\to$ M&Avg. \\
\hline
\textit{ADDA}&$89.4$&$90.1$&$89.8$\\
\hline
\textit{CoGAN}&$95.6$ &$93.1$ &$94.3$\\
\hline
\textit{UNIT}&$\bf{96.0}$ &$93.6$ &$94.8$\\
\hline
\textit{CDAN}&$93.9$ &$96.9$ &$95.4$ \\
\hline
\textit{CYCADA}&$95.6$ &$96.5$ &$\bf{96.1}$\\
\hline
\textit{Ours}&$94.1$ &$\bf{98.0}$ &$\bf{96.1}$\\
\bottomrule
\end{tabular}}
\end{center}
\setlength{\abovecaptionskip}{0pt}
\caption{Recognition accuracies ($\%$)  on handwritten digits datasets. All models utilize LeNet as base architecture.}
\label{tabmnist}
\end{table}

\begin{table}[t] \small   
\begin{center}
\setlength{\tabcolsep}{0.45mm}
{
\begin{tabular}{  l | c c  c  c  c c  |c}
\hline
Office-31&A$\to$W&D$\to$W&W$\to$D&A$\to$D&D$\to$A&W$\to$A&Avg. \\
\hline
\textit{Source Only}&$68.4$&$96.7$&$99.3$&$68.9$&$62.5$&$60.7$&$76.1$\\
\hline
\textit{TCA}&$72.7$ &$96.7$ &$99.6$ &$74.1$ &$61.7$ &$60.9$ &$77.6$\\
\textit{GFK}&$72.8$ &$95.0$ &$98.2$ &$74.5$ &$63.4$ &$61.0$ &$77.5$\\
\textit{DDC}&$75.6$ &$96.0$ &$98.2$ &$76.5$ &$62.2$ &$61.5$ &$78.3$\\
\textit{DAN}&$80.5$ &$97.1$ &$99.6$ &$78.6$ &$63.6$ &$62.8$ &$80.4$\\
\textit{RTN}&$84.5$ &$96.8$ &$99.4$ &$77.5$ &$66.2$ &$64.8$ &$81.6$\\
\textit{DANN}&$82.0$ &$96.9$ &$99.1$ &$79.7$ &$68.2$ &$67.4$ &$82.2$\\
\textit{ADDA}&$86.2$ &$96.2$ &$98.4$ &$77.8$ &$69.5$ &$68.9$ &$82.9$\\
\textit{JAN}&$85.4$ &$97.4$ &$99.8$ &$84.7$ &$68.6$ &$70.0$ &$84.3$\\
\textit{MADA}&$90.0$ &$97.4$ &$99.6$ &$87.8$ &$70.3$ &$66.4$ &$85.2$\\
\textit{SAFN}&$88.8$ &$98.4$ &$99.8$ &$87.7$ &$69.8$ &$69.7$ &$85.7$\\
\textit{GTA}&$89.5$ &$97.9$ &$99.8$ &$87.7$ &$72.8$ &$71.4$ &$86.5$\\
\textit{MCD}&$88.6$ &$98.5$ &$\bf{100.0}$ &$92.2$ &$69.5$ &$69.7$ &$86.5$\\
\textit{iCAN}&$92.5$ &$\bf{98.8}$ &$\bf{100.0}$ &$90.1$ &$72.1$ &$69.9$ &$87.2$\\
\textit{CDAN}&$94.1$ &$98.6$ &$\bf{100.0}$ &$92.9$ &$71.0$ &$69.3$ &$87.7$\\
\textit{TADA}&$94.3$ &$98.7$ &$99.8$ &$91.6$ &$72.9$ &$73.0$ &$88.4$\\
\textit{SymNet}&$90.8$ &$\bf{98.8}$ &$\bf{100.0}$ &$\bf{93.9}$ &$\bf{74.6}$ &$72.5$ &$88.4$\\
\hline
\textit{Ours}&$\bf{95.2}$ &$98.6$ &$\bf{100.0}$ &$91.7$ &$74.5$ &$\bf{73.7}$ &$\bf{89.0}$\\
\hline
\end{tabular}
}
\end{center}
\setlength{\abovecaptionskip}{0pt}
\caption{ Recognition accuracies ($\%$) on the Office31 dataset.
All models utilize ResNet-50 as base architecture.
}
\label{tab1}
\end{table}

\begin{table}[t]\small
\begin{center}
\setlength{\tabcolsep}{0.8mm}
{
\begin{tabular}{  l | c c  c  c  c c  |c}
\toprule
ImageCLEF-DA&I$\to$P&P$\to$I&I$\to$C&C$\to$I&C$\to$P&P$\to$C&Avg. \\
\hline
\textit{Source Only}&$74.8$&$83.9$&$91.5$&$78.0$&$65.5$&$91.2$&$80.7$\\
\hline
\textit{DAN} &$74.5$ &$82.2$ &$92.8$ &$86.3$ &$69.2$ &$89.8$ &$82.5$\\
\textit{RTN}&$75.6$ &$86.8$ &$95.3$ &$86.9$ &$72.7$ &$92.2$ &$84.9$\\
\textit{DANN}&$75.0$ &$86.0$ &$96.2$ &$87.0$ &$74.3$ &$91.5$ &$85.0$\\
\textit{JAN}&$76.8$ &$88.0$ &$94.7$ &$89.5$ &$74.2$ &$91.7$ &$85.8$\\
\textit{MADA}&$75.0$ &$87.9$ &$96.0$ &$88.8$ &$75.2$ &$92.2$ &$85.8$\\
\textit{iCAN} &$\bf{79.5}$ &$89.7$ &$94.7$ &$89.9$ &$\bf{78.5}$ &$92.0$ &$87.4$\\
\textit{CDAN}&$77.7$ &$90.7$ &$\bf{97.7}$ &$\bf{91.3}$ &$74.2$ &${94.3}$ &$87.7$\\
\textit{SAFN}&$78.0$ &$91.7$ &$96.2$ &$91.1$ &$77.0$ &$94.7$ &$88.1$\\
\hline
\textit{Ours}&$78.3$&$\bf{91.3}$&$96.7$&$90.5$&$78.1$&$\bf{96.2}$&$\bf{88.5}$\\
\bottomrule
\end{tabular}}
\end{center}
\setlength{\abovecaptionskip}{0pt}
\caption{Recognition accuracies ($\%$) on ImageCLEF-DA.
All models utilize ResNet-50 as base architecture.
}
\label{tab2}
\end{table}

\begin{table*}[t]\small    
\begin{center}
\setlength{\tabcolsep}{1.0mm}{
\begin{tabular}{  l | c c  c  c  c c c c  c  c  c c  |c}
\toprule
OfficeHome&Ar$\to$Cl&Ar$\to$Pr&Ar$\to$Rw&Cl$\to$Ar&Cl$\to$Pr&Cl$\to$Rw&Pr$\to$Ar&Pr$\to$Cl&Pr$\to$Rw&Rw$\to$Ar&Rw$\to$Cl&Rw $\to$Pr&Avg. \\
\hline
\textit{ResNet-50}&$34.9$&$50.0$&$58.0$&$37.4$&$41.9$&$46.2$&$38.5$&$31.2$&$60.4$&$53.9$&$41.2$&$59.9$&$46.1$\\
\hline
\textit{DAN}&$43.6$ &$57.0$ &$67.9$ &$45.8$ &$56.5$ &$60.4$ &$44.0$&$43.6$ &$67.7$ &$63.1$ &$51.5$ &$74.3$ &$56.3$\\
\textit{DANN} &$45.6$ &$59.3$ &$70.1$ &$47.0$ &$58.5$ &$60.9$ &$46.1$&$43.7$ &$68.5$ &$63.2$ &$51.8$ &$76.8$ &$57.6$\\
\textit{JAN} &$45.9$ &$61.2$ &$68.9$ &$50.4$ &$59.7$ &$61.0$ &$45.8$&$43.4$ &$70.3$ &$63.9$ &$52.4$ &$76.8$ &$58.3$\\
\textit{CDAN}&$50.7$ &$70.6$ &$76.0$ &$57.6$ &$70.0$ &$70.0$ &$57.4$&$50.9$ &$77.3$ &$70.9$ &$56.7$ &$81.6$ &$65.8$\\
\textit{SAFN}&$52.0$ &$71.7$ &$76.3$ &$\bf{64.2}$ &$69.9$ &$71.9$ &$63.7$&$51.4$ &$77.1$ &$70.9$ &$57.1$ &$81.5$ &$67.3$\\
\textit{TADA}&$53.1$ &$72.3$ &$77.2$ &$59.1$ &$71.2$ &$72.1$ &$59.7$&$53.1$ &$78.4$ &$72.4$ &$60.0$ &$82.9$ &$67.6$\\
\textit{SymNet}&$47.7$ &$72.9$ &$78.5$ &$\bf{64.2}$ &$71.3$ &$\bf{74.2}$ &$\bf{64.2}$&$48.8$ &$79.5$ &$\bf{74.5}$ &$52.6$ &$82.7$ &$67.6$\\
\hline
\textit{Ours}&$\bf{55.5}$&$\bf{73.5}$&$\bf{78.7}$&$60.7$&$\bf{74.1}$&$73.1$&$59.5$&$\bf{55.0}$&$\bf{80.4}$&$72.4$&$\bf{60.3}$&$\bf{84.3}$&$\bf{68.9}$\\
\bottomrule
\end{tabular}}
\end{center}
\setlength{\abovecaptionskip}{0pt}
\captionsetup{justification=centering}
\caption{Recognition accuracy ($\%$)  on Office-Home dataset. All models utilize ResNet-50 as base architecture.}
\label{tab3}
\end{table*}

\section{Discussion}
\textbf{Ablation Study.}
The ablation analysis results under different model variants with some loss removed are presented in Table~\ref{tab4}. The baseline of \textit{ResNet-50} denotes that only the source classifier based on cross-entropy loss is trained.
\textit{DANN} is another baseline, in which the cross-entropy loss and domain alignment are taken into account, and the performance is increased from 74.3\% to 82.1\%. Besides,
we additionally optimize the entropy minimization loss of target samples over their feature extractors and denote them as \textit{ResNet-50~(Em)} and \textit{DANN~(Em)}, respectively. They two can be regarded as another baselines compared with our method.

Our model benefits from both the novel re-weighted domain adaptation and class-level alignment cross domains.  To investigate how different components in our
model, we add every item alternately. We can verify the items from two aspects. On the one hand, we do not re-weight the adversarial loss and the \textit{DANN~(Em)} can be regarded as the baseline. Then we add the cross-domain weight ${{\mathcal{H}}_p}$ into the adversarial loss, the training setting of which is denoted as "\textit{DANN~(Em}$+{{\mathcal{H}}_p}$)", the performance is increased from 87.2\% to 88.8\% after adding re-weight item ${{\mathcal{H}}_p}$. On the other, we add the metric triplet loss $\mathcal{L}_{tri}$ into the \textit{DANN} baseline, the training setting of which is denoted as "\textit{DANN~(Em}~$+\mathcal{L}_{tri}$)" and the performance is increased from 87.2\% to 89.5\%. Furthermore, the item of "\textit{DANN~(Em}~$+\mathcal{L}_{tri}$)" can be regarded as another baseline compared with "\textit{Ours}" to prove the effectiveness of re-weight item ${{\mathcal{H}}_p}$ and the performance is increased to 90.2\%.
Additionally, because we also take some pseudo target labels into consideration, so for validating the triple loss with target samples, we have experimentally demonstrated the effectiveness of pseudo labels in metric loss. The performance is decreased from 89.5\% to 88.6\% after removing the target pseudo labels, \textit{i.e.}, "\textit{DANN~(Em}~$+{\mathcal{L}_{tri-s}}$)".

\begin{table}\small
\begin{center}
\setlength{\tabcolsep}{1.0mm}{
\begin{tabular}{  l | c c  c c |c}
\toprule
Office-31&A $\to$ W&W $\to$ D&A $\to$ D&W $\to$ A&Avg. \\
\hline
\textit{ResNet-50}&$68.4$&$99.3$&$68.9$&$60.7$&$74.3$\\
\hline
\textit{ResNet-50~(Em)}&$89.3$&$\bf{100.0}$&$89.2$&$69.0$&$86.9$\\
\Xhline{1.2pt}
\textit{DANN}&$82.0$ &$99.1$ &$79.7$&$67.4$ &$82.1$ \\
\hline
\textit{DANN~(Em)}&$89.8$ &$\bf{100.0}$ &$90.1$&$69.0$ &$87.2$ \\
\hline
\textit{DANN~(Em}$+{{\mathcal{H}}_p}$)&$92.3$ &$100.0$ &$91.1$&$71.9$ &$88.8$ \\
\Xhline{1.2pt}
\textit{DANN~(Em}$+\mathcal{L}_{tri-s}$)&$92.4$ &$99.7$ &$90.3$&$71.9$ &$88.6$ \\
\hline
\textit{DANN~(Em}$+{\mathcal{L}_{tri}}$)&$93.8$ &$99.8$ &$\bf{91.7}$&$72.6$ &$89.5$ \\
\hline
\textit{Ours}&$\bf{95.2}$ &$\bf{100.0}$ &$\bf{91.7}$&$\bf{73.7}$&$\bf{90.2}$ \\
\bottomrule
\end{tabular}}
\end{center}
\setlength{\abovecaptionskip}{0pt}
\caption{Ablation study on the Office-31 dataset.}
\label{tab4}
	\vspace{-0.12in}
\end{table}

\begin{figure}[t]
\begin{center}
  \includegraphics[width=0.95\linewidth]{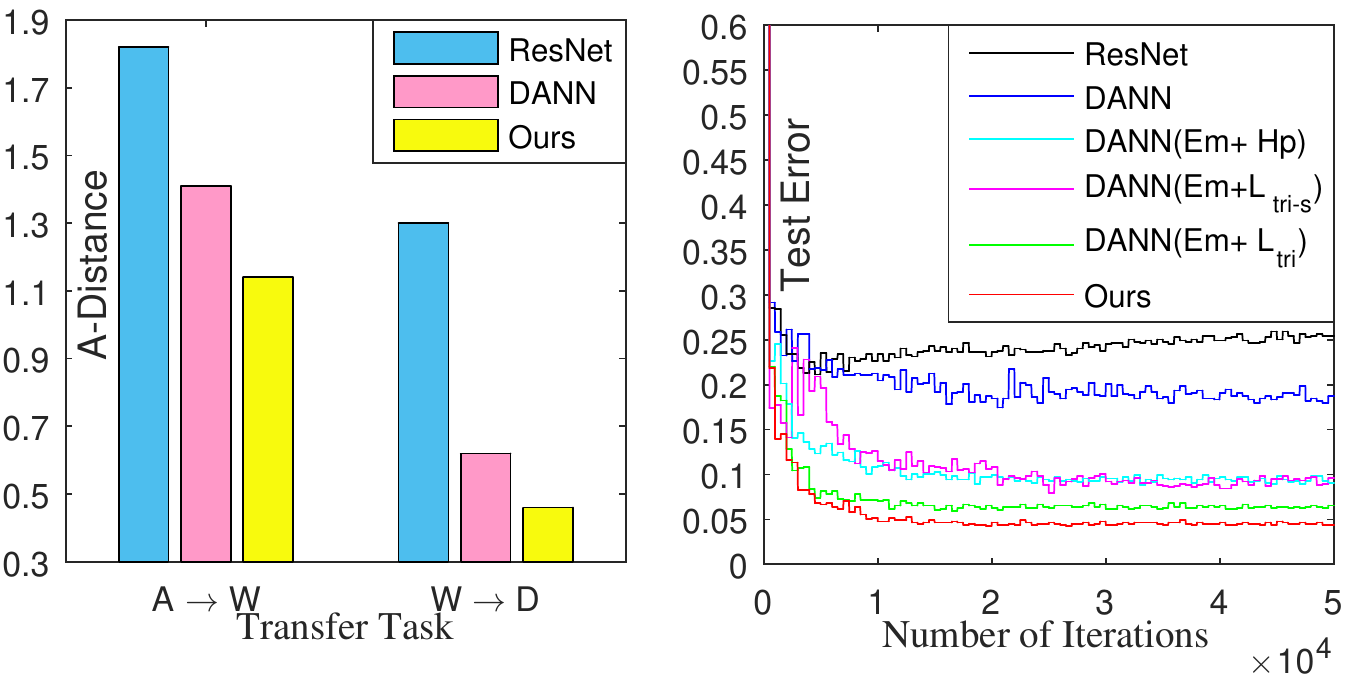}
\end{center}
\setlength{\abovecaptionskip}{0pt}
   \caption{Illustration of model analysis: (a) Quantitative distribution discrepancy measured by $\mathcal{A}$-distance after domain adaptation. (b) Convergence on the test errors of different models.}
   	\vspace{-0.12in}
   \label{fig5}
\end{figure}

\textbf{Quantitative Distribution Discrepancy}.
$\mathcal{A}$-distance \cite{Ben2010A} which jointly formulates source and target risk, are used to measure the distribution discrepancy after domain adaptation.
$\mathcal{A}$-distance is defined as $d_\mathcal{A}=2(1-2\epsilon)$, where $\epsilon$ is the classification error of a binary domain classifier (\textit{e.g.}, SVM) for discriminating the source and target domains. Therefore, with the  increasing discrepancy between two domains, the error $\epsilon$ becomes smaller.
Figure \ref{fig5}~(a) shows $\mathcal{A}$-distance on different tasks by using different models. From Figure \ref{fig5}~(a), it is obvious that a large $\mathcal{A}$-distance denotes a large domain discrepancy.
The distribution discrepancy analysis based on $\mathcal{A}$-distance in Office-31 dataset on tasks \textbf{A} $\to$ \textbf{W} and \textbf{W} $\to$ \textbf{D} is conducted by using \textit{ResNet}, \textit{DANN} and our complete model, respectively.

We can observe that $\mathcal{A}$-distance between domains after using our model is smaller than that of other two baselines, which suggests that our model is more effective in reducing the domain discrepancy gap. By comparing the distribution discrepancy between \textbf{A} $\to$ \textbf{W} and \textbf{W} $\to$ \textbf{D}, obviously, \textbf{W} $\to$ \textbf{D} has a much smaller $\mathcal{A}$-distance than \textbf{A} $\to$ \textbf{W}. From the classification accuracy in Table \ref{tab1}, the recognition rate of \textbf{W} $\to$ \textbf{D} is 100.0\%, which is higher than \textbf{A} $\to$ \textbf{W} (95.2\%). Therefore, the reliability of $\mathcal{A}$-distance is demonstrated.

\begin{figure}[t]
\begin{center}
  \includegraphics[width=0.95\linewidth]{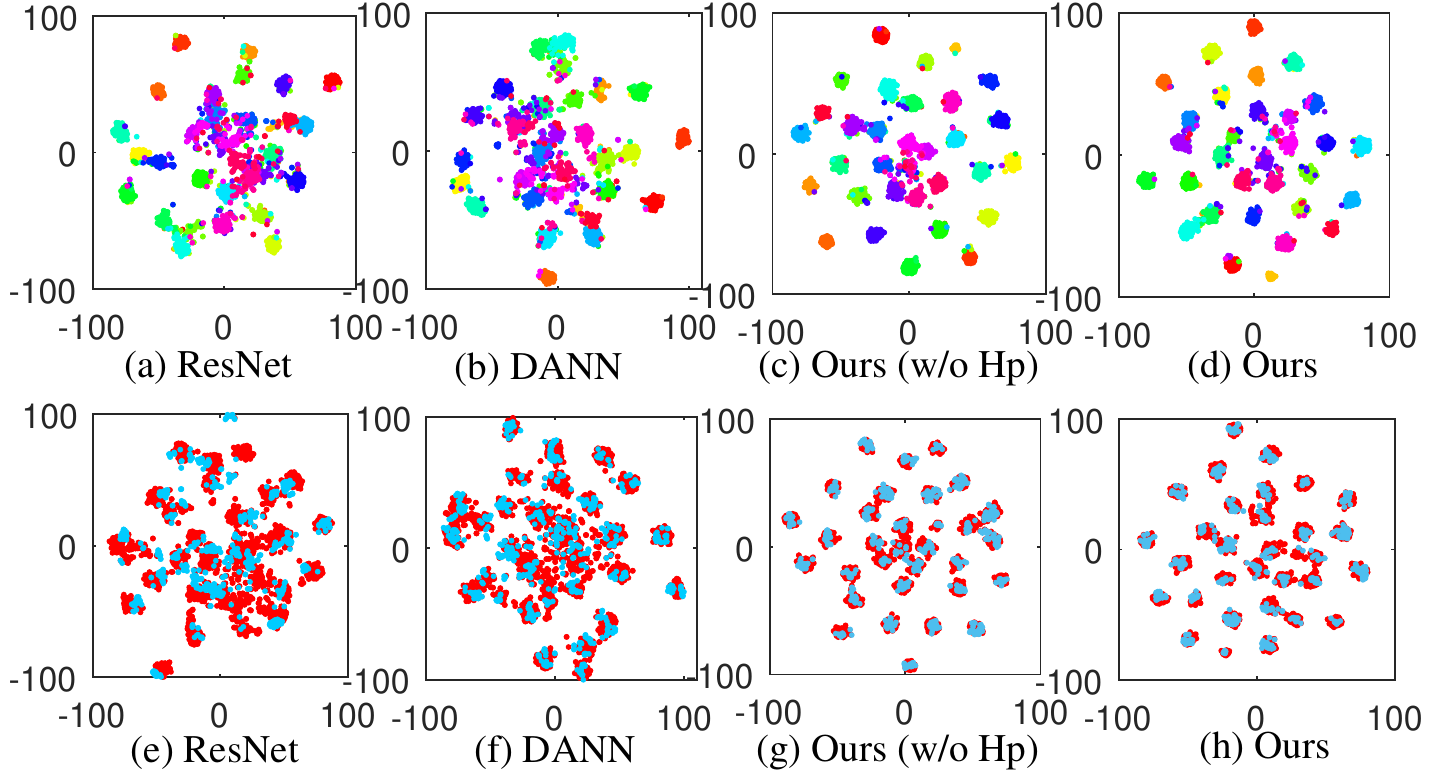}
\end{center}
  \setlength{\abovecaptionskip}{0pt}
    \caption{ Feature visualization with t-SNE algorithm.
   }
	\vspace{-0.12in}
   \label{fig7}
\end{figure}

\textbf{Convergence.}
Figure \ref{fig5} (b) shows the convergence of~\textit{ResNet}, \textit{DANN}, our baseline method only with re-weight item, \textit{i.e.}, \textit{DANN~(Em}+$\mathcal{H}_p$),  our baseline method with only source labels for triplet loss,  \textit{i.e.}, \textit{DANN~(Em}+$\mathcal{L}_{tri-s} $), our baseline method with only triplet loss,  \textit{i.e.}, \textit{DANN~(Em}+$\mathcal{L}_{tri}$), and our complete model~\textit{Ours}, respectively. We choose the task \textbf{A} $\to$ \textbf{W} in Office-31 dataset as an example and the test errors of different methods with the increasing number of iterations are shown in Figure \ref{fig5} (b).

\textbf{Feature Visualization.} 
We visualize the domain invariant features learned by \textit{ResNet}, \textit{DANN}, \textit{Ours}~($w/o~{{\mathcal{H}}_p}$), and our complete model for further validating the effectiveness. For feature visualization, t-SNE visualization method  is employed on the source domain and target domain in the \textbf{A} $\to$ \textbf{W} task from Office-31 dataset. The results of feature visualization for \textit{ResNet}~(traditional CNN), \textit{DANN}~(with adversarial learning), \textit{Ours}~($w/o~{{\mathcal{H}}_p}$)~(\textit{i.e.}, our model without re-weight), and our complete model are illustrated in Figure \ref{fig7}.

Note that Figure \ref{fig7} (a)-(d) represent the results of source features from 31 classes marked in different colors, from which we observe that \textit{Ours}~($w/o~{{\mathcal{H}}_p}$) and our model can reserve better discrimination than other two baselines as the two consider the discriminative power.
The features of two domains are visualized in Figure \ref{fig7} (e)-(h). It is obvious that the features learned by \textit{ResNet} across source and target domains can not be well aligned, without considering the feature distribution discrepancy. In \textit{DANN}, by aligning the domain distribution, the distribution discrepancy of learned features between two domains can be improved. However, the class discrepancy of features from \textit{DANN} is not improved, as \textit{DANN} does not take the class level distribution into account. In our method, compared with \textit{Ours}~($w/o~{{\mathcal{H}}_p}$), it can alleviate domain discrepancy to some extent by our re-weight. From the classification accuracies in Table \ref{tab4}, \textit{Ours}~(95.2\%) is a little better than \textit{Ours}~($w/o~{{\mathcal{H}}_p}$)~(93.8\%).
From the results, the features learned by our model can be well aligned between two domains, but reserve more class discrimination including intra-class compactness and inter-class separability.

\section{Conclusion}
To promote positive transfer and combat negative transfer in DA problem, we propose a self-adaptive re-weighted adversarial approach that tries to enhance domain alignment from the perspective of conditional distribution.
For alleviating the domain bias issue, on one hand, considering that not all images are equally transferable in domain adaptation network and some images are more transferable than others, we propose a method to reduce domain-level discrepancy by re-weighting the transferable samples. Our method reduces the weights of the adversarial loss for aligned features while increasing the adversarial forces for those poorly aligned adaptively. On the other, triplet loss is employed on the confusing domain to ensure the distance of the intra-class sample pairs closer than the inter-class pairs to achieve class-level alignment.
Therefore, the high accurate pseudo-labeled target samples and semantic alignment can be captured simultaneously in this co-training process. The experimental results verify that the proposed model outperforms state-of-the-arts in various UDA tasks.

\section*{Acknowledgements}
This work was supported by the National Science Fund of China under Grants (61771079),
Chongqing Youth Talent Program, and the Fundamental Research Funds of Chongqing (No. cstc2018jcyjAX0250).

\bibliographystyle{named}
\bibliography{ijcai20}

\end{document}